\documentclass[conference]{IEEEtran}
\IEEEoverridecommandlockouts

\usepackage{cite}
\usepackage{amsmath,amssymb,amsfonts}
\usepackage{algorithmic}
\usepackage{graphicx}
\usepackage{textcomp}
\usepackage{xcolor}
\usepackage{subfig}
\def\BibTeX{{\rm B\kern-.05em{\sc i\kern-.025em b}\kern-.08em
    T\kern-.1667em\lower.7ex\hbox{E}\kern-.125emX}}
\begin{document}

\title{Graph-Based Pseudo-multimodal Contrastive Learning for 12-Lead ECG Representations\\}

\author{\IEEEauthorblockN{1\textsuperscript{st} Mengyu Wang}
\IEEEauthorblockA{\textit{Graduate School of Engineering Science} \\
\textit{Yokohama National University}\\
Yokohama, Japan}
\and
\IEEEauthorblockN{2\textsuperscript{nd} Kozo Okada}
\IEEEauthorblockA{\textit{Division of Cardiology} \\
\textit{Yokohama City University Medical Center}\\
Yokohama, Japan}
\and
\IEEEauthorblockN{3\textsuperscript{rd} Takafumi Goto}
\IEEEauthorblockA{\textit{Technology and Innovation Department} \\
\textit{Fukuda Denshi Co.,Ltd}\\
Tokyo, Japan}
\and
\IEEEauthorblockN{4\textsuperscript{th} Natsuko Jinba}
\IEEEauthorblockA{\textit{Technology and Innovation Department} \\
\textit{Fukuda Denshi Co.,Ltd}\\
Tokyo, Japan}
\and
\IEEEauthorblockN{5\textsuperscript{th} Hiroki Yamaya}
\IEEEauthorblockA{\textit{Technology and Innovation Department} \\
\textit{Fukuda Denshi Co.,Ltd}\\
Tokyo, Japan}
\and
\IEEEauthorblockN{6\textsuperscript{th} Kiyoshi Hibi}
\IEEEauthorblockA{\textit{Division of Cardiology} \\
\textit{Yokohama City University Medical Center}\\
Yokohama, Japan}
\and
\IEEEauthorblockN{7\textsuperscript{th} Tomoki Hamagami}
\IEEEauthorblockA{\textit{Faculty of Engineering} \\
\textit{Yokohama National University}\\
Yokohama, Japan}
}

\maketitle

\begin{abstract}
12-lead electrocardiogram (ECG) is a standard, non-invasive examination widely used for diagnosing coronary artery disease, where clinical interpretation relies on comparing waveform patterns across multiple leads. 
However, most existing ECG analysis methods focus on single-lead signals or treat each lead independently, and typically process ECG signals as one-dimensional time-series data using CNNs or RNNs.
While effective in modeling local waveform changes, such approaches have difficulty capturing inter-lead dependency and global waveform patterns essential for clinical diagnosis.

To address this limitation, we propose a graph-based pseudo-multimodal contrastive learning framework called Graph-CMMC. 
ECG waveforms are transformed into Gramian Angular Difference Field (GADF) images to construct complementary representations of the same cardiac activity, enabling a pseudo-multimodal learning setting. 
Using all 12 leads, Graph-CMMC aligns waveform and GADF representations in a self-supervised manner, while a graph-based relational module is employed to model inter-lead dependency and enforce structural consistency across leads during contrastive learning. 

Experimental results on a multi-label coronary artery occlusion classification task demonstrate that the proposed framework achieves competitive performance compared to supervised learning methods.
These results further suggest the effectiveness of using GADF as a complementary representation and incorporating explicit graph-based modeling of inter-lead dependency for learning robust 12-lead ECG representations.
\end{abstract}

\begin{IEEEkeywords}
12-lead ECG, Contrastive learning, Pseudo-multimodal representation, Graph-based modeling, Self-supervised learning
\end{IEEEkeywords}

\section{Introduction}
12-lead electrocardiogram (ECG) is a standard and non-invasive examination that records cardiac electrical activity from 12 different viewpoints and plays an essential role in the diagnosis of coronary artery disease. 
By comparing waveform patterns across multiple leads, clinicians can infer the location and severity of cardiac abnormalities. 
As a result, accurate ECG interpretation relies not only on temporal waveform morphology within each lead, but also on the relationships among leads.

In recent years, AI-based ECG analysis methods have been actively studied to support clinical diagnosis. 
Most existing approaches process ECG signals as one-dimensional time-series data using convolutional neural networks (CNNs) or recurrent neural networks (RNNs). 
While these methods are effective in capturing local temporal patterns, they often focus on single-lead signals or treat each lead independently.

Moreover, a representation gap exists between current AI-based ECG analysis methods and actual clinical diagnostic practice. 
Physicians do not interpret ECG signals solely as sequential numerical values. 
Instead, they visually inspect global waveform shapes, rhythm patterns, and their consistency across leads. 
Time-series-based processing, which emphasizes point-wise numerical processing, may therefore be insufficient to capture the structural characteristics underlying multi-lead ECG signals. 
This mismatch can limit the robustness and reliability of learned representations, particularly in settings where labeled data are limited.

To address these challenges, we propose a graph-based pseudo-multimodal contrastive learning framework called Graph-CMMC, for representation learning from 12-lead ECG signals. 
In the proposed framework, ECG waveforms are transformed into Gramian Angular Difference Field (GADF)~\cite{gadf} images to construct complementary representations of the same cardiac activity. 
While waveform signals preserve fine-grained temporal morphology, GADF representations emphasize global correlation structures over time. 
This pseudo-multimodal formulation allows complementary representations to be obtained from the same ECG recordings, without requiring any additional data.

Based on these representations, Graph-CMMC employs contrastive learning in a self-supervised manner to align waveform and GADF representations across all 12 leads. 
In addition, a graph-based relational module is introduced to explicitly model inter-lead dependency and enforce structural consistency during representation learning. 

The proposed method is evaluated on a multi-label coronary artery occlusion classification task using real-world 12-lead ECG recordings to assess its effectiveness in learning robust representations.

In this work, we make the following contributions.

\begin{itemize}
  \item We propose Graph-CMMC, a graph-based pseudo-multimodal contrastive learning framework for representation learning from 12-lead ECG signals, which leverages complementary waveform and GADF representations during pretraining, rather than as multimodal inputs in the downstream task.

  \item We introduce a learnable graph structure to explicitly model inter-lead dependency and incorporate this dependency directly into the representation learning process.

  \item Through experiments and analyses, we show that the proposed framework not only enables the encoder to learn more stable representations, but also learns transferable inter-lead graph structures that contribute to improved downstream performance.
  \end{itemize}

\section{Related Works}
\subsection{Time-series-based ECG analysis}
Deep learning has been widely applied to automated ECG analysis, particularly through time-series models such as CNNs and RNNs, including LSTM~\cite{lstm} and GRU variants~\cite{gru}. 
These approaches model ECG signals as one-dimensional temporal sequences and have shown strong performance in detecting cardiac abnormalities such as arrhythmias and atrial fibrillation~\cite{deepthanrule}. 

However, most time-series-based methods process ECG signals per-lead and treat each lead independently. 
Such designs make it difficult to explicitly capture inter-lead dependency, which plays an important role in the interpretation of 12-lead ECG recordings.

In addition to direct time-series modeling, several studies have explored time-series-to-image transformations for ECG analysis, such as Gramian Angular Field and its variant GADF.
These approaches convert ECG waveforms into two-dimensional images that emphasize waveform morphology and temporal correlation patterns, and have been applied to tasks such as myocardial infarction detection.
Yousuf et al.~\cite{MI_gaf} applied GAF-based representations to single-lead ECG signals and demonstrated their effectiveness for myocardial infarction classification.
Similarly, Zhang et al.~\cite{MI_PCANet} employed GADF-based representations with a PCANet architecture, showing robustness to noise and stable feature extraction.
However, similar to many time-series-based approaches, these studies mainly focused on single-lead ECG analysis, typically using only Lead II, which limits their applicability to multi-lead ECG analysis.

\subsection{Multi-lead ECG modeling}
Recent studies have explored multi-lead ECG modeling to better capture the complementary information across different leads.
For example, Zhou et al.~\cite{multilead} proposed a hierarchical convolutional framework with lead encoder attention to exploit multi-lead features for ECG classification.
Such approaches explicitly consider the contributions of individual leads and have demonstrated improved performance over single-lead models.

These studies highlight the importance of multi-lead modeling in ECG analysis.
However, most existing multi-lead ECG models are designed in a supervised learning setting and model inter-lead relationships within task-specific architectures.

\subsection{Self-supervised ECG representation learning}
Self-supervised learning has gained increasing attention in ECG analysis due to its ability to learn meaningful representations from unlabeled signals.
Existing self-supervised ECG representation learning methods can be broadly categorized into reconstruction-based approaches and contrastive learning-based approaches.

Reconstruction-based methods typically adopt masked modeling strategies to recover missing signal segments.
A representative method is ST-MEM, proposed by Na et al.~\cite{stmem}, which applies masked modeling to 1D ECG time-series signals.
ST-MEM introduces lead-wise patch masking and reconstructs masked signals using a shared Transformer encoder with sinusoidal positional embeddings and lead embeddings.
This design enables the model to capture intra-lead temporal structures while leveraging shared representations across multiple leads.

Despite its effectiveness, reconstruction-based methods such as ST-MEM have several limitations.
Operating directly on raw 1D signals makes it difficult to capture global morphological patterns, such as ST-segment elevation or QRS deformation, which involve long-range temporal structures.
In addition, reconstruction objectives focus on accurately recovering the input signal, making them sensitive to local fluctuations and baseline drift, which can limit their ability to learn robust and discriminative representations for downstream diagnostic tasks.

In contrast, contrastive learning-based methods aim to learn invariant representations by aligning positive pairs while separating negative samples.
By directly optimizing representation similarity and separation, contrastive learning is well suited for learning transferable ECG representations.

CLOCS is a family of contrastive learning methods for ECG analysis, proposed by Kiyasseh et al.~\cite{clocs}.
Among its components, Contrastive Multi-lead Coding (CMLC) focuses on exploiting inter-lead relationships.
CMLC treats temporally aligned ECG signals from different leads as positive pairs, based on the observation that different leads correspond to distinct projections of the same underlying cardiac electrical activity.
By contrasting lead-wise views of the same ECG recording, CMLC captures spatial invariance across leads and incorporates inter-lead information into the learned representations.

However, in existing contrastive ECG representation learning methods such as CMLC, inter-lead relationships are primarily used for defining positive pairs, rather than being explicitly modeled as structured relational constraints during representation learning.
Moreover, these methods typically operate within a single representation space and do not explicitly ensure that inter-lead relationships remain consistent across different ECG representations.

\section{Proposed Method}
\begin{figure}[ht]
\centering
  \includegraphics[scale = 0.3]{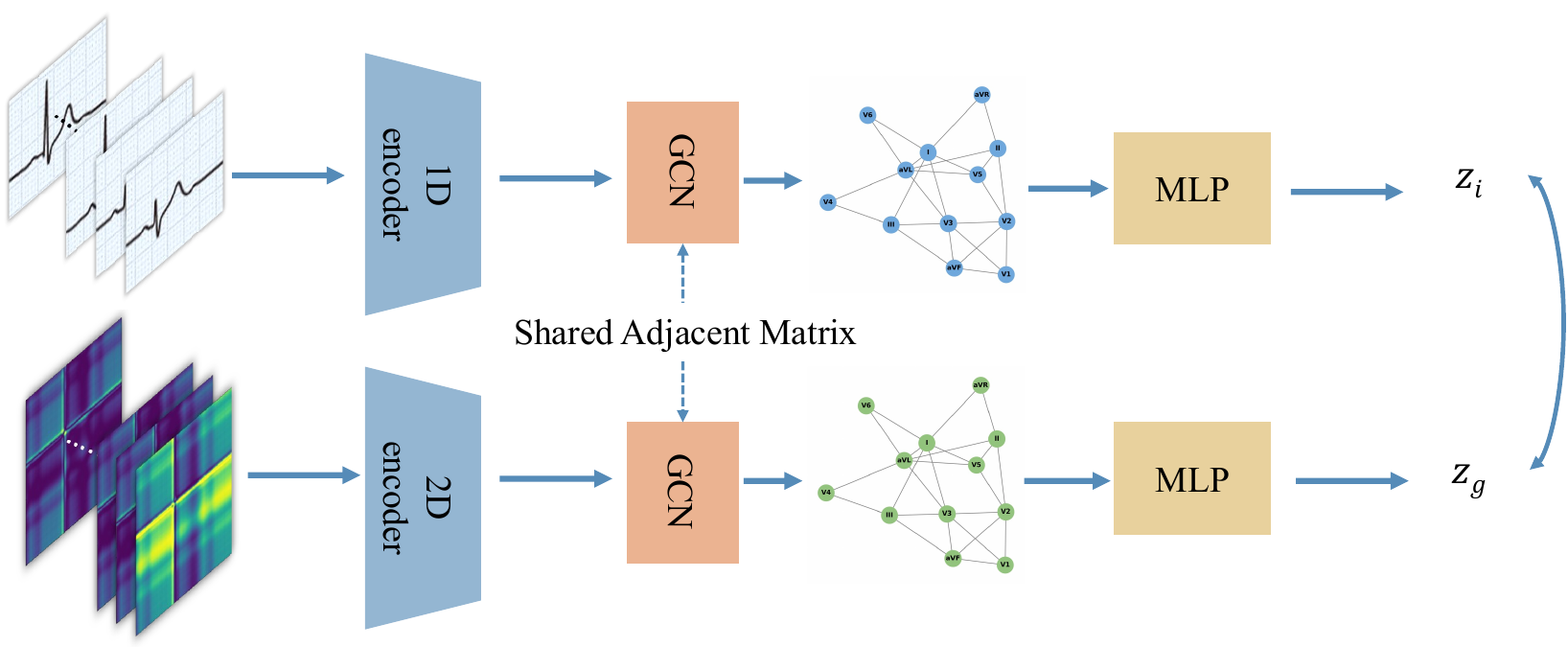}
  \caption{An overview of Graph-CMMC}
\label{fig:graph-cmmc}
\end{figure}

To address these limitations, we propose Graph-CMMC (Graph-based Cross-Modal Matching Contrastive Learning), a self-supervised framework designed for 12-lead ECG signals.
An overview of Graph-CMMC is illustrated in Fig.~\ref{fig:graph-cmmc}.

Given a 12-lead ECG recording, the framework first constructs a pseudo-multimodal representation by encoding the same cardiac activity in two forms: raw waveform signals and GADF images.
These representations are processed by modality-specific encoders implemented as lightweight 2-layer CNNs, while inter-lead relationships are explicitly modeled using a shared graph structure.
Finally, a contrastive learning objective is applied to align waveform and GADF representations in a self-supervised manner. 

\subsection{Pseudo-multimodal}
To bridge the representation gap between time-series-based ECG modeling and clinical diagnostic practice, Graph-CMMC adopts a pseudo-multimodal representation strategy.
Here, pseudo-multimodal representation refers to constructing multiple complementary representations from the same 12-lead ECG recording, where GADF images provide a two-dimensional view emphasizing global waveform structures that may be less explicit in raw waveform signals.

\subsubsection{GADF Transform}
The GADF transform maps temporal relationships in ECG signals into a spatial representation by encoding pairwise angular differences over time.
Compared to raw waveform signals, this two-dimensional representation emphasizes global waveform morphology and temporal correlation patterns that are less explicit in time-series form and are closely related to the visual inspection process used by clinicians.

\begin{figure}[ht]
\centering
  \includegraphics[scale = 0.2]{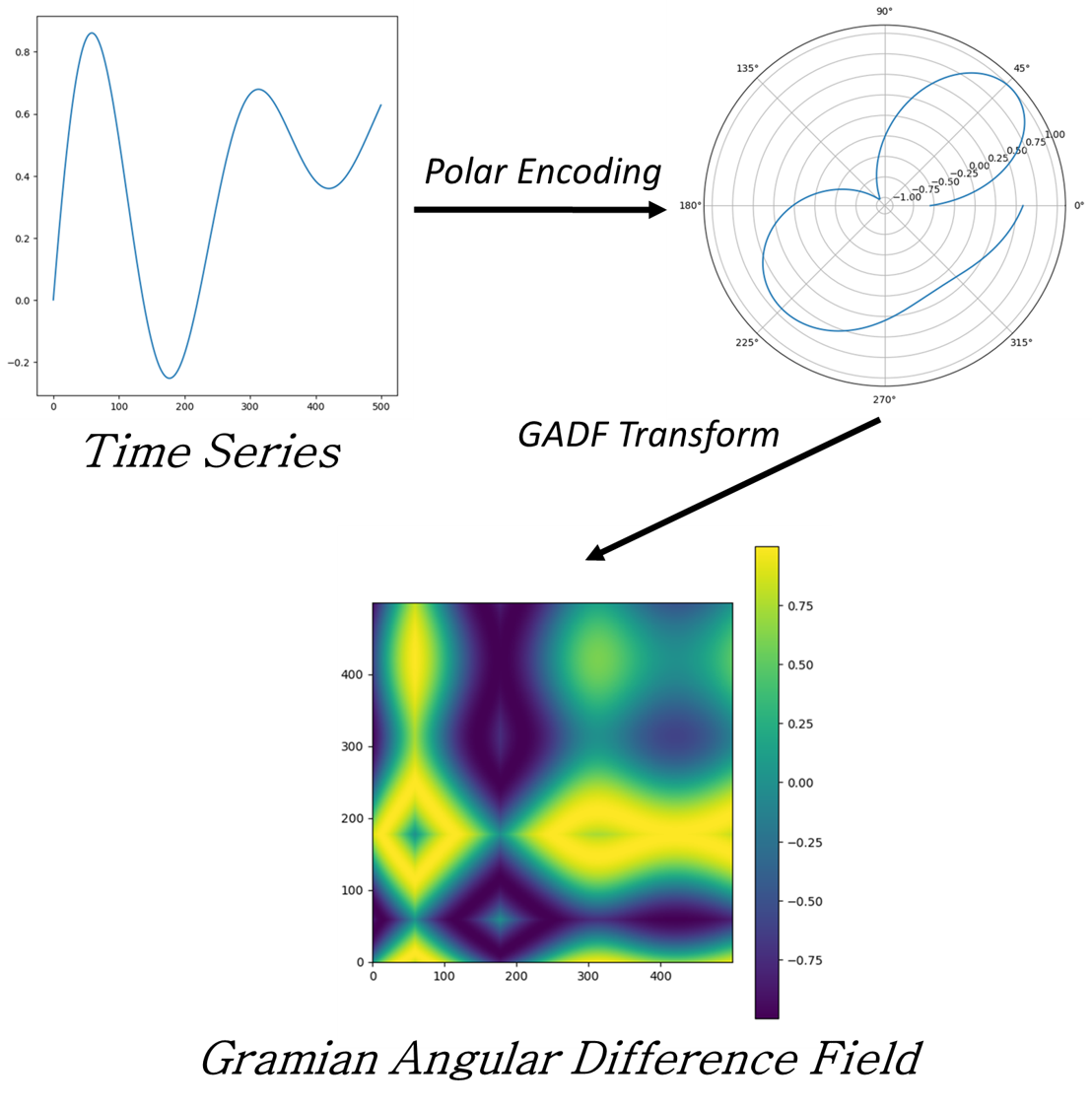}
  \caption{An overview of GADF Transform}
\label{fig:gadf}
\end{figure}

As shown in Fig.~\ref{fig:gadf}, the GADF transformation involves three steps.
First, the ECG signal is normalized to the range $[-1, 1]$.
\begin{equation}
x_t^{\prime} = \frac{x_t - \min(\boldsymbol{x})}{\max(\boldsymbol{x}) - \min(\boldsymbol{x})} \times 2 - 1 .
\label{eq:scaling}
\end{equation}
Second, the normalized signal is mapped to angular values.
\begin{equation}
\phi_t = \arccos(x_t^{\prime}) .
\label{eq:toAngle}
\end{equation}
Finally, the GADF image is constructed by computing the sine of pairwise angular differences.
\begin{equation}
\label{eq:toGADF}
\centering
\textbf{GADF}=\left[\begin{array}{ccc}\sin \left(\phi_1-\phi_1\right) & \cdots & \sin \left(\phi_1-\phi_T\right) \\ \sin \left(\phi_2-\phi_1\right) & \cdots & \sin \left(\phi_2-\phi_T\right) \\ \vdots & \ddots & \vdots \\ \sin \left(\phi_T-\phi_1\right) & \cdots & \sin \left(\phi_T-\phi_T\right)\end{array}\right]
\end{equation}
where $T$ denotes the length of the time series, and the generated GADF image is resized to $64 \times 64$ in this study.

Through this process, a one-dimensional ECG waveform is converted into a structured, pattern-like two-dimensional image.
In this study, GADF is introduced as a complementary representation during pretraining to encourage the model to capture such global structural information.

It should be noted that the pseudo-multimodal design in Graph-CMMC is introduced specifically for self-supervised pretraining, rather than for increasing the complexity of the downstream model.
By aligning waveform-based and GADF-based representations during pretraining, the encoder is encouraged to capture both local temporal patterns and global structural characteristics.
In the downstream task, only the waveform-based encoder is used in order to evaluate the quality and transferability of the learned representations under a simple and consistent setting.

\subsection{Graph-based Module}
To explicitly capture inter-lead dependency in 12-lead ECG signals, Graph-CMMC incorporates a graph-based relational module.
Each ECG lead is treated as a node in a graph, and lead-wise relationships are represented using an adjacency matrix that encodes inter-lead dependency.

The graph-based module operates on lead-level representations extracted by the modality-specific 2-layer CNN encoders.
After feature extraction, a graph convolution layer is applied to these lead-level representations to enable relational message passing across leads.
This design enables information exchange across leads and supports modeling inter-lead dependency in multi-lead ECG signals.

In the proposed framework, the adjacency matrix is learned through backpropagation in the waveform-based branch rather than the GADF-based branch.
At each training step, the updated adjacency matrix is shared with the GADF-based branch and used for graph convolution.
The GADF-based branch does not independently optimize the adjacency matrix, but instead utilizes the shared relational structure learned from waveform representations.
This sharing refers to the use of a common inter-lead dependency structure, rather than sharing network parameters across modalities.

This design choice is motivated by the fact that waveform representations preserve direct temporal variations across leads, providing a more reliable basis for modeling lead-wise dependency.
GADF representations are derived views that emphasize global structural patterns and are therefore used as complementary representations, rather than as the primary source for learning relational structure.

By modeling inter-lead relationships through a shared graph structure, Graph-CMMC encourages waveform-based and GADF-based representations to respect consistent lead-wise dependency patterns during representation learning.

\subsection{Contrastive Learning}
Graph-CMMC adopts a SimCLR-style \cite{simclr} contrastive learning framework to align waveform-based and GADF-based representations.
The objective is to encourage representations derived from the same ECG recording to be close in the embedding space, while separating representations from different recordings.

The resulting representations from the waveform and GADF branches are aggregated and projected into a shared embedding space using a lightweight MLP-based projection head.
These projected representations are then used for contrastive learning.

For a given ECG recording, the waveform-based representation and the corresponding GADF-based representation are treated as a positive pair.
Representations from different ECG recordings within the same mini-batch are treated as negative samples.

The contrastive objective is defined using the Normalized Temperature-scaled Cross-entropy (NT-Xent) loss.
Given a positive pair $(\mathbf{z}_i, \mathbf{z}_g)$, where $\mathbf{z}_i$ and $\mathbf{z}_g$ denote the projected representations from the waveform-based and GADF-based branches of the same ECG recording, respectively, the loss is defined as

\begin{equation}
\mathcal{L}_{i,g} = -\log \frac{\exp(\mathrm{sim}(\mathbf{z}_i, \mathbf{z}_g)/\tau)}
{\sum_{k \neq i} \exp(\mathrm{sim}(\mathbf{z}_i, \mathbf{z}_k)/\tau)}
\end{equation}
where $\mathrm{sim}(\cdot,\cdot)$ denotes cosine similarity and $\tau$ is a temperature parameter.
In practice, a symmetric formulation is adopted, where the contrastive loss is computed in both directions, i.e., waveform-to-GADF and GADF-to-waveform, and the results are averaged.
This loss encourages alignment between corresponding waveform and GADF representations while separating non-corresponding pairs within the batch.

By optimizing this objective, Graph-CMMC aligns waveform and GADF representations while preserving inter-lead dependency through the shared graph structure.

\section{Experiment}
\subsection{Dataset}
The 12-lead ECG dataset used in this study was collected from real clinical recordings provided by Yokohama City University Medical Center.

The dataset consists of ECG recordings from 1,068 patients.
Each ECG sample is annotated by physicians with coronary artery occlusion information, where the culprit vessel was determined based on coronary angiography.

The original dataset contains 38,516 ECG segments, each with a temporal length of 10,000 points.
For experimental evaluation, ECG segments were extracted using a fixed-length temporal window of 2,000 points, corresponding to approximately 4 cardiac cycles, to ensure consistent input length across samples.
To maintain a balanced label distribution and reduce the impact of class imbalance, a total of 3,641 ECG samples were selected for this study, with approximately 1,000 samples for each coronary artery label.

The downstream task is formulated as a multi-label classification problem.
Each ECG sample may be associated with one or multiple occluded coronary arteries.
In this study, four major coronary arteries are considered: 
left anterior descending artery (LAD), left circumflex artery (LCX), right coronary artery (RCA), and left main trunk (LMT).

\subsection{Experimental Settings}
The dataset is split into training, validation, and test sets with a ratio of 6:2:2.
Patient-level separation is not enforced during data splitting, and as multiple ECG samples are available for some patients, ECG data from the same patient may appear across different subsets.
To ensure fair comparison, the random seed is fixed so that all self-supervised methods are trained on the same data distribution.

The experiments are conducted in two stages: self-supervised pretraining and downstream multi-label classification.

\subsubsection{Self-supervised Pretraining}
Self-supervised pretraining is performed using the entire training set without using any label information.

During pretraining, models are trained for 200 epochs with a batch size of 8.
The Adam optimizer is used with an initial learning rate of $1 \times 10^{-4}$.
All models are trained using the same network architecture and optimization settings to ensure a fair comparison.

\subsubsection{Downstream Classification}
For downstream classification, the pretrained waveform-based encoder is extracted and used as a fixed feature extractor.
A multi-layer perceptron (MLP) classifier is then attached to the encoder output for multi-label classification.

The MLP classifier consists of 2 fully connected layers with batch normalization, GELU activation, and dropout.
During downstream training, the pretrained encoder is frozen and only the classification head is optimized.

The classification head is trained for 50 epochs with a batch size of 8 using the Adam optimizer.
The learning rate is set to $1 \times 10^{-3}$.
Early stopping based on validation performance is applied to prevent overfitting.

\subsubsection{Evaluation Metrics}
Model performance is evaluated using multiple metrics suitable for multi-label classification.

\begin{itemize}
	\item{Macro F1-score}: The F1-score is computed independently for each label and then averaged across all labels.
	This metric reflects balanced performance among classes and is insensitive to label imbalance.

	\item{Subset accuracy}: A strict evaluation metric that counts a prediction as correct only when the predicted label set exactly matches the ground-truth label set for a given sample.
	
	\item{Overall accuracy}: Overall accuracy is computed as $1 - Hamming Loss$, where Hamming Loss measures the fraction of incorrectly predicted labels over all labels, providing a label-wise accuracy measure.
	
	\item{Jaccard index}: the Jaccard index (sample-based Intersection over Union) is used to evaluate the overlap between predicted and ground-truth label sets for each sample.

\end{itemize}

These metrics are computed on the test set.
All experiments are repeated three times, and the results are reported as the average over these runs.

\subsubsection{Comparison methods}
To evaluate the effectiveness of the proposed Graph-CMMC, several self-supervised and supervised methods are selected for comparison.
All comparison methods use the same dataset split, encoder architecture, and downstream classification head to ensure a fair evaluation.

\begin{itemize}
	\item \textbf{Supervised baseline}: A fully supervised model with the same 2-layer 1D CNN encoder and MLP classifier as Graph-CMMC, trained end-to-end using labeled data.
		
	\item \textbf{CMLC}~\cite{clocs}: As shown in Fig.~\ref{fig:cmlc}, CMLC performs contrastive learning by treating ECG signals from different leads of the same patient as positive pairs, while signals from other patients serve as negative samples.
	
\begin{figure}[ht]
\centering
  \includegraphics[scale = 0.25]{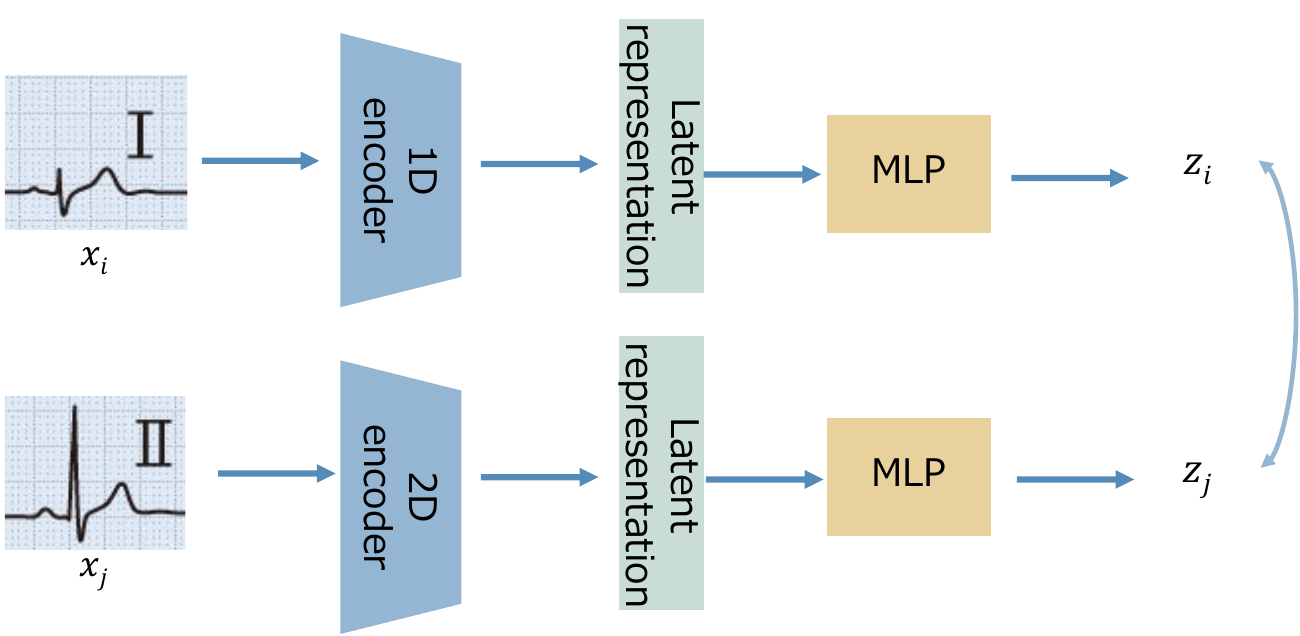}
  \caption{An overview of CMLC}
\label{fig:cmlc}
\end{figure}
			
	\item \textbf{CMMC}: As shown in Fig.~\ref{fig:cmmc}, CMMC performs contrastive learning by aligning waveform-based and GADF-based representations of the same ECG recording as positive pairs, while representations from different recordings are treated as negative samples.
	Unlike Graph-CMMC, CMMC considers each lead independently and does not model inter-lead relationships using graph structures.
	
\begin{figure}[ht]
\centering
  \includegraphics[scale = 0.25]{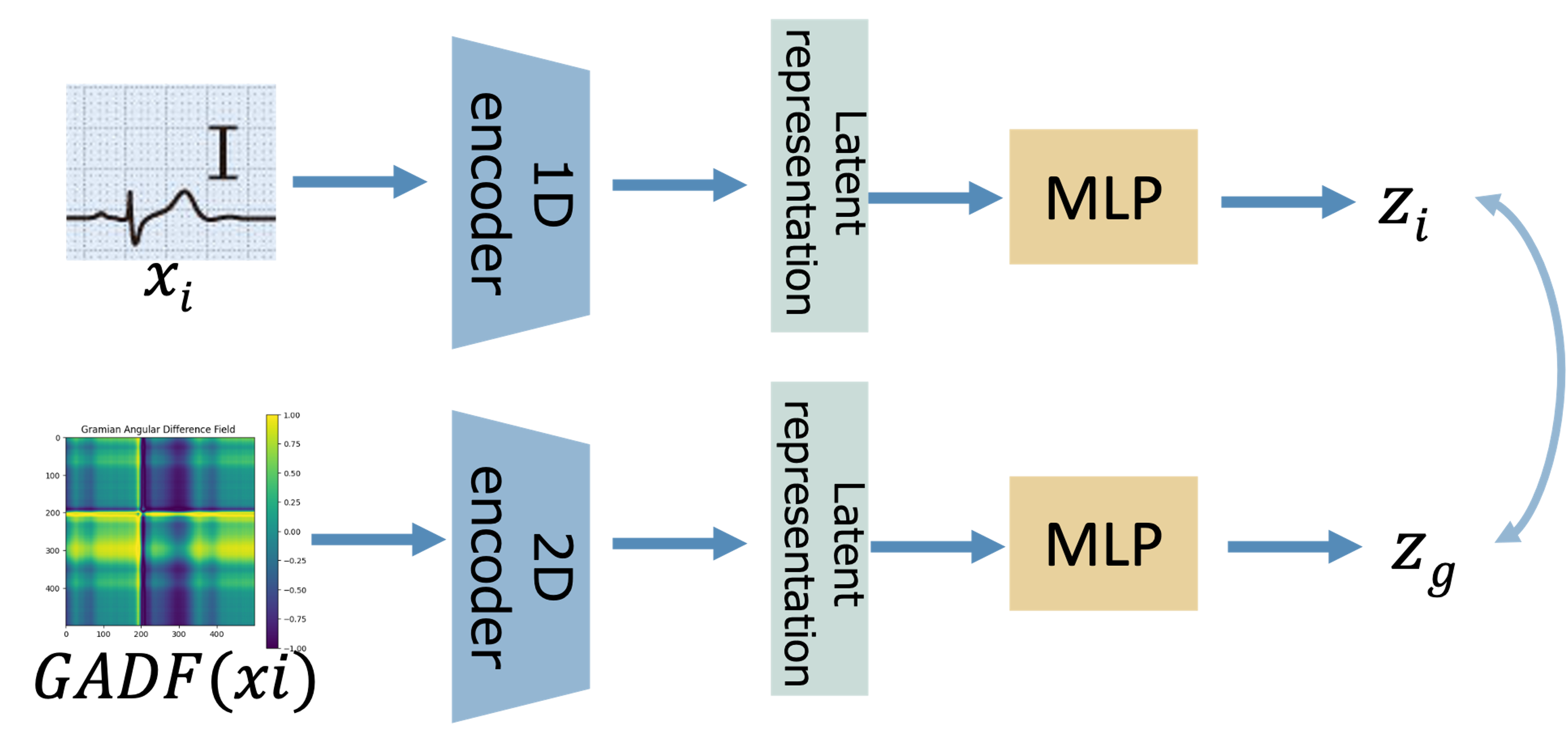}
  \caption{An overview of CMMC}
\label{fig:cmmc}
\end{figure}

	\item \textbf{CMMC+CMLC}: The overview of CMMC+CMLC is shown in Fig.~\ref{fig:cmmccmlc}.
	This method combines CMMC and CMLC by jointly applying cross-modal contrastive learning and multi-lead contrastive learning.
\begin{figure}[ht]
\centering
  \includegraphics[scale = 0.35]{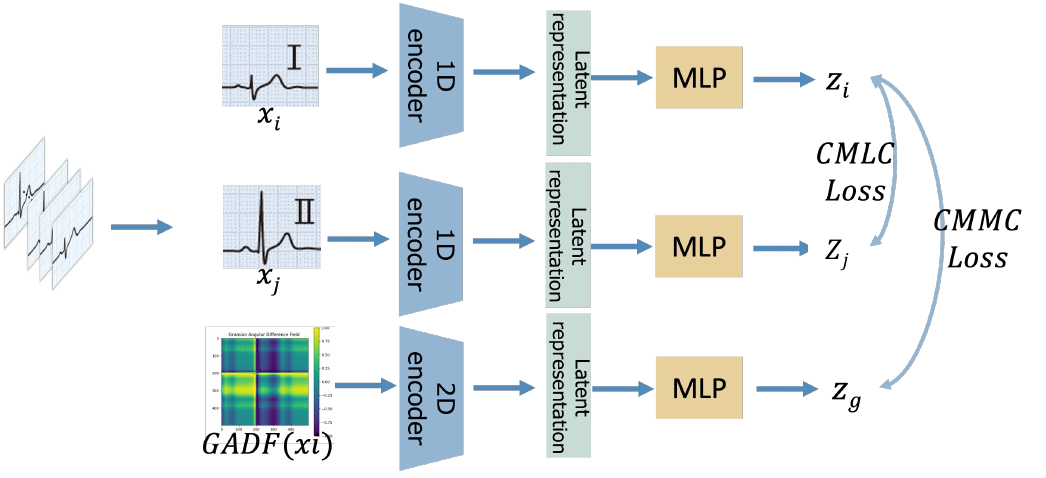}
  \caption{An overview of CMMC+CMLC}
\label{fig:cmmccmlc}
\end{figure}

\end{itemize}

\subsection{Results}
For the proposed Graph-CMMC, two downstream evaluation settings are considered to examine how the inter-lead dependency learned during self-supervised pretraining is transferred to the downstream task.

In Graph-CMMC (d1), the pretrained waveform-based encoder is followed by an MLP classifier for multi-label classification.
In Graph-CMMC (d2), the inter-lead adjacency matrix learned during pretraining is incorporated into the downstream model to perform graph-based feature aggregation before classification.
By comparing these two settings, we evaluate whether the inter-lead dependency learned in the self-supervised stage remains effective after encoder transfer.

Table~\ref{tab:results} summarizes the test performance of all methods in terms of Macro F1-score, Subset Accuracy, Overall Accuracy, and Jaccard index.
The fully supervised baseline achieves the highest performance and serves as an upper-bound reference.

Among the self-supervised methods, Graph-CMMC achieves the best performance across all evaluation metrics.
Graph-CMMC (d2) outperforms Graph-CMMC (d1), indicating that incorporating the learned inter-lead graph structure at the downstream stage provides additional gains beyond encoder-only transfer.

Regarding the comparison methods, CMLC shows lower performance across all metrics, suggesting that contrastive learning based only on inter-lead pairing is insufficient for this multi-label coronary artery occlusion task.
CMMC improves over CMLC, indicating that aligning waveform and GADF representations leads to more informative features.
However, its performance remains below Graph-CMMC, which models inter-lead dependency through a learnable graph structure.
The combined method, CMLC+CMMC, does not outperform CMMC alone, suggesting that combining multi-lead and cross-modal contrastive objectives does not necessarily lead to additive improvements.

Table~\ref{tab:per_class} reports class-wise F1-scores for each coronary artery.
Graph-CMMC (d2) achieves higher F1-scores than Graph-CMMC (d1) for all four classes (LAD, LCX, LMT, and RCA), and outperforms the other self-supervised baselines across classes.
These results indicate that the learned inter-lead modeling benefits multiple coronary regions rather than improving only a specific class.

\begin{table}[ht]
\centering
\caption{Experimental results}
\label{tab:results}
\begin{tabular}{lcccc}
\hline
Method 
& Macro F1 
& \begin{tabular}[c]{@{}c@{}}Subset \\Accuracy\end{tabular} 
& \begin{tabular}[c]{@{}c@{}}Overall \\Accuracy\end{tabular} 
& \begin{tabular}[c]{@{}c@{}}Jaccard\end{tabular} \\
\hline
Supervised baseline & 0.822 & 0.771 & 0.903 & 0.786 \\ \hline
CMLC                & 0.563 & 0.432 & 0.797 & 0.515 \\
CMLC+CMMC           & 0.703 & 0.593 & 0.846 & 0.628 \\
CMMC                & 0.729 & 0.627 & 0.857 & 0.665 \\
Graph-CMMC (d1)     & 0.746 & 0.670 & 0.855 & 0.715 \\
Graph-CMMC (d2)     & \textbf{0.776} & \textbf{0.711} & \textbf{0.873} & \textbf{0.752} \\
\hline
\end{tabular}
\end{table}

\begin{table}[ht]
\centering
\caption{Per-class F1-score}
\label{tab:per_class}
\begin{tabular}{lcccc}
\hline
Model & LAD & LCX & LMT & RCA \\
\hline
Supervised baseline & 0.781 & 0.809 & 0.936 & 0.763 \\
\hline
CMLC                & 0.328 & 0.526 & 0.769 & 0.629 \\
CMLC+CMMC           & 0.680 & 0.602 & 0.834 & 0.696 \\
CMMC                & 0.726 & 0.607 & 0.859 & 0.723 \\
Graph-CMMC (d1)     & 0.739 & 0.697 & 0.861 & 0.686 \\
Graph-CMMC (d2)     & \textbf{0.758} & \textbf{0.720} & \textbf{0.880} & \textbf{0.745} \\
\hline
\end{tabular}
\end{table}
\subsection{Discussion}
\subsubsection{Feature space analysis of Inter-Lead Representations}
To investigate where the performance differences among methods originate, we analyzed the inter-lead structure learned by each self-supervised approach.
For each ECG sample, lead-wise embeddings were extracted from the waveform-based encoder
\(
h_i \in \mathbb{R}^D,\ i = 1,\ldots,12,
\)
and Euclidean distances were computed for all unique lead pairs, resulting in 66 pairwise distances for the 12-lead ECG.
The distribution of these distances was summarized using four statistics: mean, standard deviation, minimum, and maximum.

Table~\ref{tab:distance} reports the inter-lead distance statistics for each method.
CMLC exhibits a large standard deviation together with extreme minimum and maximum distances, indicating that both overly similar and overly separated lead pairs are present, resulting in an unstable inter-lead structure.
CMMC shows a more balanced distance distribution, with moderate mean distance and reduced variance.
However, when CMLC is combined with CMMC, the mean distance increases and the variance remains relatively large, suggesting that the combined objective does not improve the stability of the learned structure.

In contrast, Graph-CMMC achieves the smallest mean distance and the lowest standard deviation among all methods.
This indicates that the inter-lead structure is learned in a more compact and stable manner, without collapsing into trivial representations or excessively separating specific leads.

\begin{figure}[ht]
\centering
  \includegraphics[scale = 0.07]{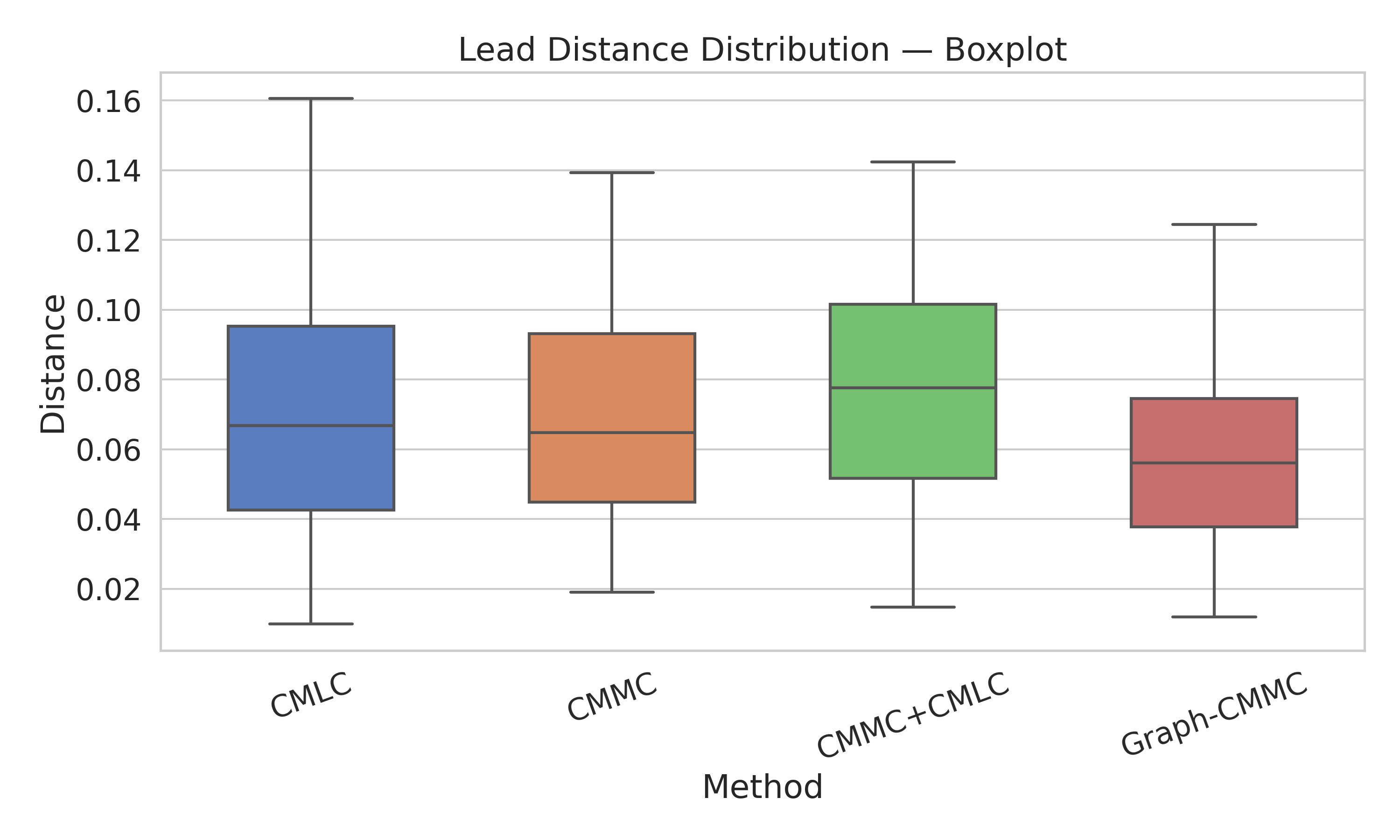}
  \caption{Boxplot comparison of inter-lead distance distributions for different methods}
\label{fig:distance}
\end{figure}

Fig.~\ref{fig:distance} provides a distribution-level view of inter-lead distances and supports the observations in Table~\ref{tab:distance}.
Graph-CMMC shows a narrower interquartile range and fewer extreme values compared with the other methods, indicating more consistent inter-lead relationships across samples.

A possible interpretation is related to how each contrastive objective treats lead-specific information.
CMLC encourages representations from different leads to become similar, which may suppress lead-specific characteristics.
CMMC does not explicitly enforce such alignment, resulting in a more averaged distance distribution.
When these objectives are combined, conflicting optimization pressures may arise and destabilize the feature space.
In contrast, Graph-CMMC explicitly models inter-lead dependency through a graph structure, which is consistent with the observed compact and stable representations.

\begin{table}[t]
\centering
\caption{Inter-lead embedding distance statistics for different methods}
\label{tab:distance}
\begin{tabular}{lcccc}
\hline
Method & Mean ($\downarrow$) & Std ($\downarrow$) & Min  & Max \\
\hline
CMMC        & 0.0659 & 0.0265 & 0.0188 & 0.1190 \\
CMLC        & 0.0648 & 0.0335 & 0.0145 & 0.1735 \\
CMMC+CMLC   & 0.0639 & \textbf{0.0264} & 0.0150 & 0.1381 \\
Graph-CMMC  & \textbf{0.0589} & 0.0265& 0.0119 & 0.1248 \\
\hline
\end{tabular}
\end{table}

\subsubsection{Visualization of learned Inter-lead dependency in Graph-CMMC}
To further examine what type of inter-lead dependency is captured by Graph-CMMC, we visualize the learned adjacency matrix and its community structure.

\begin{figure}[ht]
\centering
\subfloat[Heatmap]{
  \includegraphics[scale=0.19]{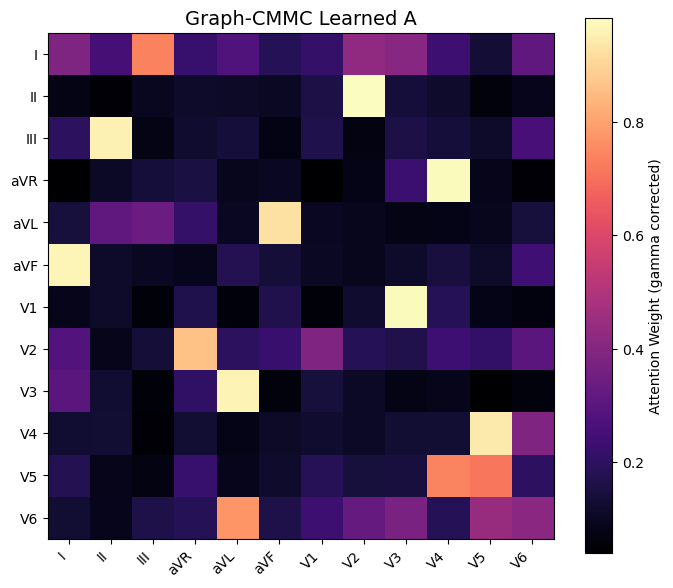}
  \label{fig:heatmap}
}
\hfill
\subfloat[Louvain clustering results ]{
  \includegraphics[scale=0.19]{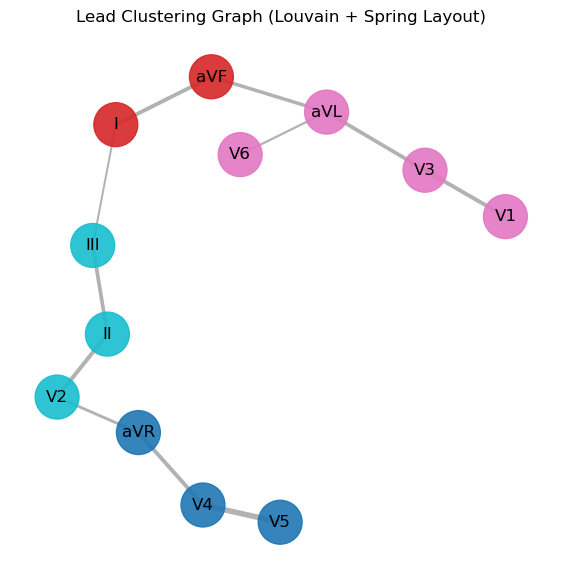}
  \label{fig:graph}
}
\caption{Visualization of the inter-lead dependency learned by Graph-CMMC.
(a) Heatmap of the learned adjacency matrix.
(b) Louvain clustering results based on the learned adjacency matrix.}
\label{fig:interlead}
\end{figure}

As shown in Fig.~\ref{fig:heatmap}, the adjacency matrix learned by Graph-CMMC is not symmetric.
This indicates that the model captures directional dependencies, where features from one lead contribute differently to the representation of another lead.
Such asymmetry is expected in 12-lead ECG signals, as each lead observes cardiac electrical activity from a different spatial orientation.
Therefore, the learned asymmetric structure reflects directional information flow among leads rather than assuming mutual equivalence.

As an example of the learned dependency structure, Fig.~\ref{fig:graph} shows the inter-lead communities obtained by applying Louvain clustering~\cite{louvain} to the learned adjacency matrix.
The 12 leads are partitioned into four communities: 
$\{\mathrm{I}, \mathrm{aVF}\}$,
$\{\mathrm{aVL}, \mathrm{V1}, \mathrm{V3}, \mathrm{V6}\}$,
$\{\mathrm{aVR}, \mathrm{V4}, \mathrm{V5}\}$,
and
$\{\mathrm{II}, \mathrm{III}, \mathrm{V2}\}$.

The resulting communities do not strictly correspond to anatomical groupings defined in clinical textbooks, and no clear consistency with clinical interpretations was observed.
This reflects a limitation of data-driven approaches, which primarily extract statistical dependencies from training data rather than explicitly encoding clinically defined structures.

Despite this limitation, the adjacency matrix learned by Graph-CMMC captures inter-lead dependencies that are informative for the downstream task.
Although the learned structure does not directly match predefined anatomical groupings, it emphasizes task-relevant relationships among leads that are beneficial for the downstream task.

\section{Conclusion}
In this study, we proposed Graph-CMMC, a self-supervised graph-based pseudo-multimodal contrastive learning framework for representation learning from 12-lead ECG signals.
The proposed method constructs complementary waveform and GADF representations from the same ECG recording and aligns them through contrastive learning, while explicitly modeling inter-lead dependency using a learnable graph structure.

By incorporating a graph-based relational module, Graph-CMMC enables structured information exchange across leads and enforces consistency of inter-lead relationships during representation learning.
Compared with existing self-supervised methods, the proposed method achieves superior performance on a multi-label coronary artery occlusion classification task, approaching the performance of fully supervised learning.
Furthermore, incorporating the learned inter-lead graph structure in the downstream stage provides additional performance gains, indicating that the dependency learned during self-supervised pretraining remains effective after encoder transfer.

Additional analyses of the learned feature space demonstrate that Graph-CMMC produces more compact and stable inter-lead representations than comparison methods.
Visualization of the learned adjacency matrix further reveals directional inter-lead dependencies that reflect task-relevant information flow among leads, although these learned structures do not strictly correspond to clinically predefined anatomical groupings.

This study highlights the effectiveness of combining pseudo-multimodal representation learning with explicit graph-based modeling for 12-lead ECG analysis.
Graph-CMMC serves as a step toward bridging the gap between data-driven ECG representation learning and the multi-lead relational reasoning employed in clinical practice.

For future work, we plan to evaluate the proposed framework on larger and more diverse ECG datasets, including experiments with patient-wise data splitting.
We will also investigate lightweight graph designs and conduct ablation studies on architectural choices such as encoder depth and the number of graph convolution layers.
In addition, different downstream strategies and incorporating domain knowledge will be explored to better understand and interpret the learned inter-lead dependency structures.

\section*{Acknowledgment}
This study protocol was reviewed and approved by the Yokohama National University Ethics Committee for Medical and Biological Research Involving Human Subjects (Approval No. Hitoi-2023-28).

\vspace{12pt}

\end{document}